\documentclass[letterpaper, 10 pt, conference]{ieeeconf}

\IEEEoverridecommandlockouts

\usepackage{amsmath} 

\usepackage{balance}
\usepackage{amssymb}
\usepackage{amsfonts}
\usepackage{enumerate}
\usepackage{tabulary}
\usepackage{multirow}
\usepackage{url}
\usepackage{cite}
\usepackage{lipsum}  
\usepackage{graphicx}
\usepackage{subfigure}
\usepackage{epstopdf}
\usepackage[misc]{ifsym}
\usepackage{color}
\usepackage{xcolor}
\usepackage{xxcolor}
\usepackage{setspace}
\usepackage[skins]{tcolorbox}
\usepackage{listings}
\usepackage{lipsum}
\usepackage{booktabs}
\usepackage[hidelinks]{hyperref}
\usepackage[lined,boxed,commentsnumbered,linesnumbered]{algorithm2e}
\usepackage{cleveref}

\usepackage{siunitx} 

\usepackage{flushend}

\definecolor{codegreen}{rgb}{0,0.6,0}
\definecolor{codepurple}{rgb}{0.58,0,0.82}
\definecolor{backcolour}{rgb}{0.95,0.95,0.92}
\lstdefinestyle{buzz}{
    backgroundcolor=\color{black!5},   
    commentstyle=\color{codegreen},
    keywordstyle=\color{blue},
    numberstyle=\tiny\color{black!30},
    stringstyle=\color{codepurple},
    basicstyle=\footnotesize\ttfamily,
    breakatwhitespace=false,         
    breaklines=true,                 
    captionpos=b,                    
    keepspaces=true,                 
    numbers=left,                    
    numbersep=5pt,                  
    showspaces=false,                
    showstringspaces=false,
    showtabs=false,                  
    tabsize=2,
}
\newcommand{\fixed}[1]{\textcolor{black}{#1}}

\usepackage[colorinlistoftodos,prependcaption,textsize=tiny]{todonotes}

\title{\LARGE \bf
Vision-Based Online Key Point Estimation of Deformable Robots
}

\author{Hehui Zheng*$^{1,2}$, Sebastian Pinzello$^{1}$, Barnabas Gavin Cangan$^{1}$, Thomas Buchner$^{1}$, 
Robert K. Katzschmann$^{1}$
\thanks{*Correspondence {\tt hehui.zheng@srl.ethz.ch}}%
\thanks{$^{1}$Soft Robotics Lab, ETH Zürich, Zürich, Switzerland;}%
\thanks{$^{2}$ETH AI Center; Zürich, Switzerland.}%
}

\begin{document}

\maketitle
\thispagestyle{empty}
\pagestyle{empty}

\begin{abstract}
The precise control of soft and continuum robots requires knowledge of their shape, which has, in contrast to classical rigid robots, infinite degrees of freedom.
To partially reconstruct the shape, proprioceptive techniques use built-in sensors resulting in inaccurate results and increased fabrication complexity. Exteroceptive methods so far rely on expensive tracking systems with reflective markers placed on all components, which are infeasible for deformable robots interacting with the environment due to marker occlusion and damage.
Here, a regression approach is presented for 3D key point estimation using a convolutional neural network. 
The proposed approach takes advantage of data-driven supervised learning and is capable of online marker-less estimation during inference. 
Two images of a robotic system are taken simultaneously at 25 Hz from two different perspectives, and are fed to the network, which returns for each pair the parameterized key point or PCC shape representations.
The proposed approach outperforms marker-less state-of-the-art methods by a maximum of 4.5\% in estimation accuracy while at the same time being more robust and requiring no prior knowledge of the shape. Online evaluations on two types of soft robotic arms and a soft robotic fish demonstrate our method's accuracy and versatility on highly deformable systems.
\fixed{\footnotemark[7]}

\end{abstract}

\footnotetext[7]{All code available on \url{https://github.com/srl-ethz/voke}.}

\section{Introduction}
\label{sec:intro}
Soft robots are experiencing a steep rise in popularity thanks to their ability to solve challenges such as compliant grasping and dexterous movement,\cite{yamanaka2020development, wang2020dual} tasks with which rigid robots typically struggle.\cite{hawkes2021hard}
To fully exploit the capabilities of soft robots, modern control approaches are needed, which typically rely on rich state feedback.
However, obtaining and accurately describing the state of a continuously deforming soft body or robot is challenging compared to the state of a rigid object or robot.
Encoders at the connecting joints of rigid robots readily provide precise state measurements, while soft robots mostly consist of elastomeric materials that deform with infinite degrees of freedom.\cite{yasa_overview_2022} It is, therefore, crucial to solve the challenge of soft robotic state estimation to exploit the full potential of the great variety of soft robots for manipulation and beyond.

To date, various estimation approaches have attempted to improve soft robotic sensing capabilities.
One type of sensing approach uses mechanical proprioception similar to classical robotic state estimation.\cite{park2019multi, wang2020mechanoreception}
A variety of sensor types such as resistive, capacitive, optical, and pneumatic transducers proprioceptively estimate the continuous deformations of soft robots.\cite{wang2018toward, hofer2021vision}
Mechanical proprioception with built-in sensors is limited by spatial resolution and increased fabrication complexity.
This limitation has lately led to an increased popularity of exteroceptive sensing approaches that are purely vision-based.\cite{bern2020soft,albeladi2021vision}

\begin{figure}
    \centering
    \includegraphics[width=0.9\linewidth]{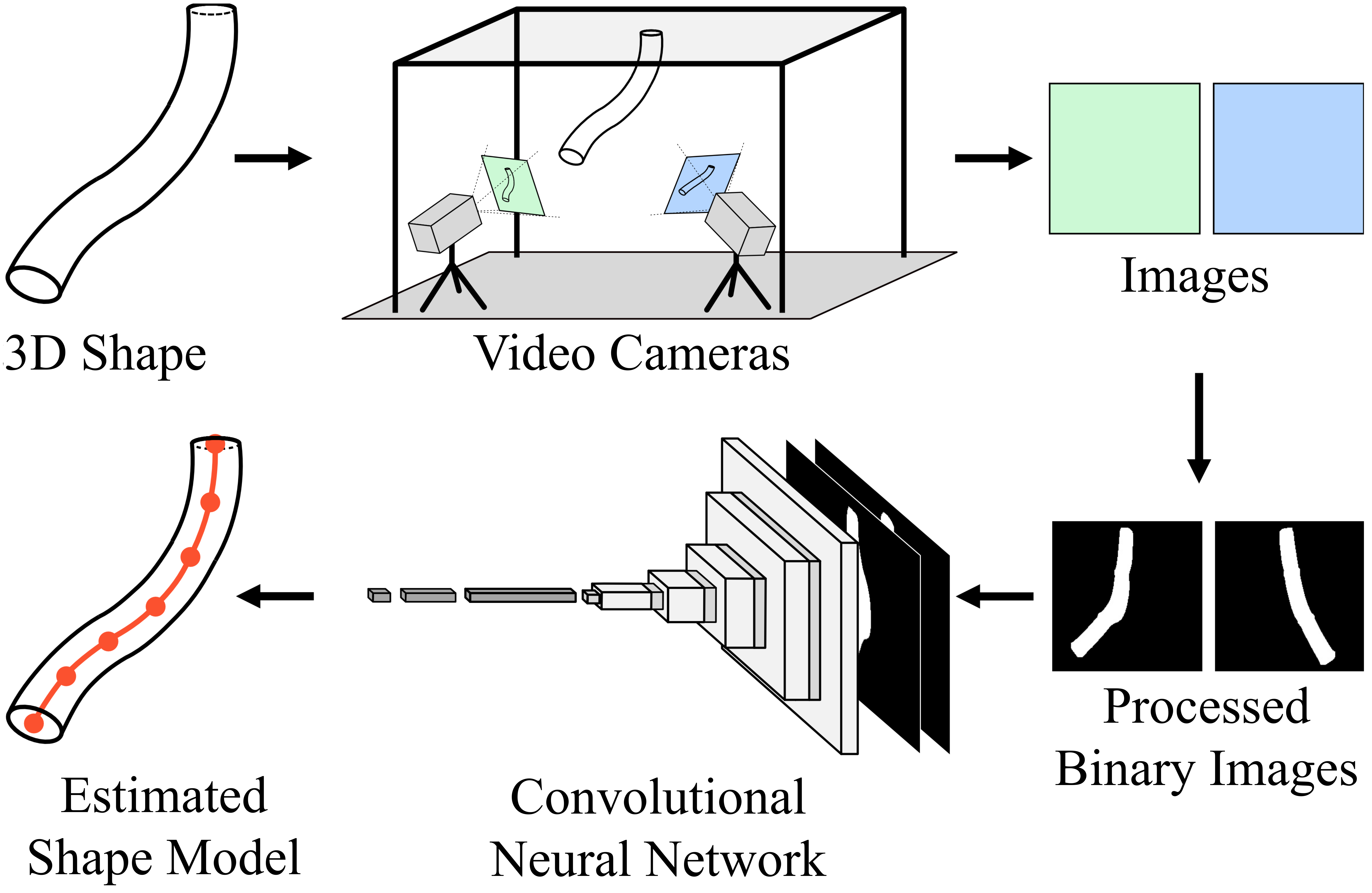}
    \caption{Diagram of the marker-less online inference pipeline for the proposed key point estimation approach VOKE. A 3D shape is captured by two RGB video cameras. Image pairs are preprocessed and run through a convolutional neural network to estimate a shape model for the 3D shape.}
    \label{fig:vise_pipeline}
\end{figure}

A widely applied vision-based method for key point estimation and tracking employs motion-capture systems.\cite{bern2020soft,katzschmann2019dynamic,marchese2014design}
These systems rely on placing reflective markers on an object and triangulating the marker position using multiple calibrated motion-tracking cameras.
Such marker-based approaches offer high temporal and spatial resolutions of the placed markers.
However, these tracking systems are usually costly and limited in use, hindering the application of marker-based robots in commercial settings where multiple robots are deployed simultaneously within clutter.\cite{patrizi2016comparison}
Motion-capture systems require the tedious placement of many markers on any object to be tracked, which is not a feasible option when working with high-dimensional robotic systems that interact with an environment.
Markers cannot be placed too close to each other, limiting the fidelity of shape reconstruction.
Furthermore, obstacles could occlude and eventually displace or destroy the markers.
Therefore, much research has gone into replacing marker-based approaches by developing marker-less sensing techniques.\cite{hofer2021vision,albeladi2021vision,hannan2005real, camarillo2008vision, croom2010visual, ceseracciu2014comparison, vandini2017unified}
Current marker-less approaches still require scenario-specific capturing setups and strict formatting requirements of the image data. 

In this work, we choose an approach towards 3D key point estimation of continuously deformable robots that leverages data-driven deep learning.
We propose VOKE, a {V}ision-based regression approach for camera-based, {O}nline 3D {K}ey Point {E}stimation using a convolutional neural network (CNN) (\textbf{Figure 1}). 
The two types of estimated parametric shape representations used in this work are the key point position and the piecewise constant curvature model.
The performance of our online estimation is compared against marker-based motion capture in the tasks of estimating the shape of soft robotic arms and soft robotic fish. Although we do not explicitly consider occlusion in the scope of this paper, we also provide preliminary evaluations of the performance with occlusion or under different unseen experimental setups.

Specifically, our work provides the following contributions:
\begin{itemize}
    \item We present a CNN-based approach using two cameras that is applicable to various soft robots.
    \item We conduct a comprehensive performance comparison of various CNN architectures, demonstrating that VGG architectures achieve the fastest inference speeds and the best estimation accuracy on soft arm datasets. 
    \item We demonstrate the online estimation capability of our system, which is a critical step towards enabling closed-loop control of soft robots. While this paper focuses on the estimation aspect, the results lay the groundwork for future research into integrating these estimations into closed-loop control systems, thereby enhancing the autonomy and responsiveness of soft robots.
\end{itemize}

Section 2 summarizes related work and Section 3 presents our methodology. In Section 4, we provide evaluation results to exhibit the performance of the proposed CNN-based approach and discuss the estimation accuracy under different model assumptions. Section 5 concludes the work and outlines directions for future research. Finally, in the Experimental Section we give the details of our tested soft robots and experimental setup.

\section{Related Works}
\label{sec:related}
Previous works demonstrate vision-based shape parameter estimation approaches for continuously deformable robots that either work only for specific setups or necessitate strict image requirements, \textit{e.g.}, exact segmentation for contour extraction.
In the following, we briefly discuss the most related works, focusing on marker-less shape estimation approaches.

Hannan and Walker use basic image processing techniques, including thresholding and image segmentation, to estimate the 2D shape of a planar, cable-actuated elephant trunk manipulator from single images.\cite{hannan2005real} 
However, their estimation results are only compared to another cable-based shape estimation technique but not with actual ground truth.
Camarillo et al. extend the computer-vision methods to 3D spline estimation of a thin continuum manipulator.\cite{camarillo2008vision}
If the precise positions of cameras are known, they could extract silhouettes from multiple cameras' views, project those silhouettes into a volumetric space to find their intersection, and fit a spline through the resulting 3D point cloud. This approach requires a strong contrast between the tracked shape and the background, as well as the absence of other objects in the field of view.

Strict requirements on the image data are also found in other works.
AlBeladi et al. rely on successful contour extraction of their soft arm to fit a geometric strain-based model to these edges.\cite{albeladi2021vision}
Croom et al. also perform edge detection, but then fit reference points to the edges by using an unsupervised learning algorithm called the self-organizing map.\cite{croom2010visual}
All of these approaches show good estimation results but require a strong contrast between the tracked object and the background.

Vandini et al. extract and join straight lines from a monoplane fluoroscopic surgical image to estimate the shape of a concentric tube robot.\cite{vandini2017unified}
By posing conditions for connecting the line features, they manage to relax image requirements and can extract curves from more unclean image data compared to the aforementioned works.\cite{vandini2017robust}
Reiter et al. take on a similar approach to ours in that they extract features from segmented binary stereo-images.\cite{reiter2012learning}
Since their feature extraction relies on the color-coded segments of their continuum robot, it does not generalize to other robots that do not have those features.

Mathis et al. created a deep learning framework based on transfer learning for marker-less pose estimation and tracking called DeepLabCut.\cite{mathis2018deeplabcut}
Their framework enables tracking of multiple visual features in unprocessed videos using only a small number of labeled frames for training.
They demonstrate their method by tracking body parts of mice and show that they achieve pixel tracking errors comparable to human-level labeling accuracy.
However, this framework by itself is restricted to pixel tracking in an image, and it cannot directly track 3D coordinates of features.

Our approach for 3D key point estimation of continuously deformable robots employs convolutional neural networks.
While there are many vision-based proprioceptive methods for soft robots using deep learning,\cite{werner2019vision, wang2020real, she2020exoskeleton} we focus on exteroceptive approaches that are simple to implement and do not add complexity to the manufacturing of the soft robots.

\section{Methods}
\label{sec:methods}
Our proposed estimation method is a learned multi-view parametric estimation from grayscale images.
First, different views of the desired object are captured by two cameras.
Then, unnecessary information in the images is removed by an image processing pipeline that transforms them into binary images.
The processed images are then fed into a convolutional neural network (CNN) trained to estimate the parameters of the shape representation.
The approach is outlined in Figure 1 and the following subsections detail the sub-components of our method.

\subsection{Image Preprocessing}
\label{sec:image_preprocessing}
The RGB images of the shape are preprocessed to facilitate the learning procedure.
The original images are converted to grayscale, cropped around the shape, and scaled to the size of 256$\times$256 pixels.
This preprocessing preserves enough information to accurately represent the shape, while keeping the number of parameters of the CNN relatively small.
A median filter with a 7$\times$7 pixel kernel size is applied to reduce noise before using adaptive thresholding to reduce the grayscale image to a binary image.\cite{opencv_library}
This step removes the background variations while preserving the shape.
In addition, the adaptive nature of the thresholding operation and the resulting binary images make our trained network function in a wide range of lighting conditions without the need for retraining.
Erosion and dilation with a 7$\times$7 pixel kernel are applied for three iterations each to remove the remaining artifacts.

\begin{figure}
    \centering
    \includegraphics[width=0.9\linewidth]{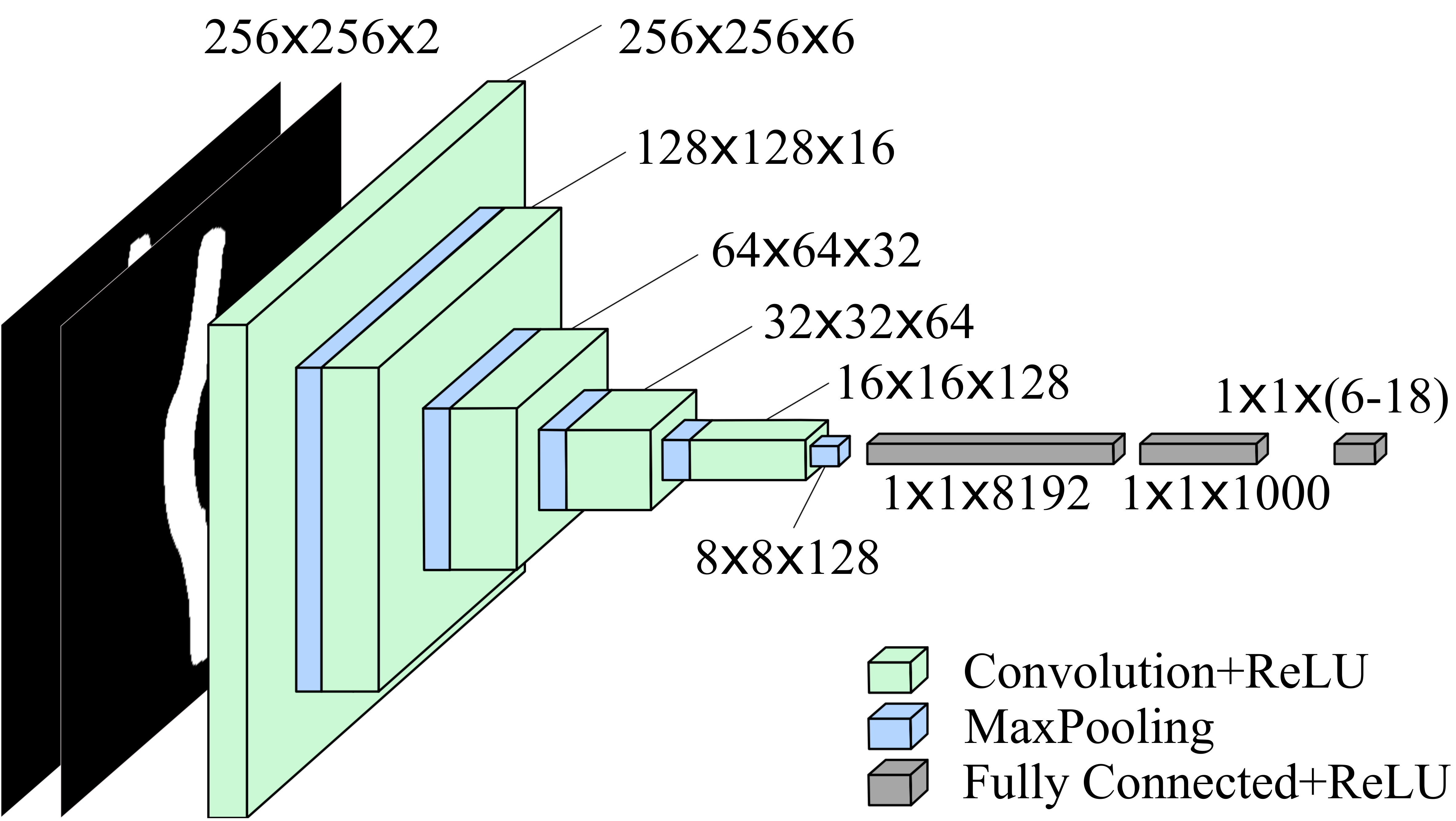}
    \caption{Truncated VGG Network Architecture: VGG-s-bn. Inputs are the preprocessed binary images from both cameras, and output sizes depend on the selected shape representation and robot (Table 1).}
    \label{fig:network_arch}
\end{figure}

\subsection{Shape Representations}
\label{sec:ground_truth}
We consider two different parametric shape representations, the point position, and piecewise constant curvature (PCC) model.
Ground truth shape is obtained using the motion capture software (\textit{Qualisys Track Manager}). Virtual coordinate frames are placed at the center of each group of motion markers to line up with the corresponding segment's centroid.
These coordinate frames allow the tracking of not only each segment's translation but also orientation.
The point representation employs only the translation and comprises the positions of the key points along the soft robots, requiring three parameters for each key point.
The PCC model,\cite{webster2010design, della2020improved} a commonly used kinematic reduction model in soft robotics,can be fitted to both the translation and orientation of the virtual coordinate frames, allowing the modeling of a continuous shape by approximating it with multiple constant-curvature sections of fixed length.
Each section is defined by two parameters, the curvature and an angle indicating the curving direction.
Hence, compared to the point representation, the PCC model requires fewer parameters, 6 instead of 9 for a three-section soft arm, while representing a long, continuously deformable shape, which is useful for model-based control purposes. 

\subsection{Network Architecture}
\label{sec:network_architecture}
In this work, we compare the performance of the following CNN architectures: VGG\cite{simonyan2014very} (VGG11, VGG11-bn with batch normalization, VGG13, VGG13-bn with batch normalization), ResNet\cite{he2016deep} (ResNet18, ResNet34, Resnet50), EfficientNet\cite{tan2019efficientnet} (EffNet-B0, EffNet-B1, EffNet-B3, EffNet-B5), as well as EfficientNetV2\cite{tan2021efficientnetv2} (EffNetV2-s, EffNetV2-m).
For all architectures, we modify the input channels to 2 and reduce the output size to 6, 9, 12, or 18, depending on the shape representation and number of sections to estimate. The output sizes can be found in \textbf{Table 1}.

To avoid overfitting with large networks, we also introduce a custom truncated network, VGG-s-bn, adapted from the VGG architecture.
Several convolutional layers are removed from the standard VGG network to further reduce the computational demand  and improve online performance.
The final soft-max layer is also removed to perform a regression instead of a classification.
The network architecture is illustrated in \textbf{Figure 2}.

The network's main elements are convolutional layers, batch normalization, rectification (ReLU) nonlinearities, and max pooling operations.
These elements are applied in the mentioned order and repeated five times before the output is fed into two fully connected layers.
All five convolutional layers have a kernel of size three, a stride of one, and a padding of one.
The number of channel dimensions is increased from 2 to 6, 16, 32, 64, and 128.
Batch normalization is applied before each convolutional layer.
Every max-pooling operation reduces its input by a factor of two, reducing the initial image size of 256$\times$256 pixels to 8$\times$8 pixels after five operations.
Hence, the input to the first fully connected layer is of size 8,192 (8$\times$8$\times$128).
A ReLU nonlinearity is applied after the first fully connected layer.
The input to the last fully connected layer has a size of 1,000, which is reduced to the output size of 6, 9, 12, or 18 (Table 1).

\subsection{Camera Realignment}
\label{sec:camera_realignment}
No explicit camera calibration is needed and the camera configuration is implicitly learned by the CNN. This design choice limits the cameras' positions to be fixed relative to the soft robot during the data collection.
To alleviate the need for retraining, fiducial markers (\textit{AprilTags}) are attached to the robot's base.\cite{olson2011apriltag}
The camera's translation and rotation relative to the base can be extracted from the image of the fiducial markers.
Our realignment utility for the camera pose compares the camera's current and previously saved positions relative to the fiducial markers. With this utility, users can set up the RGB cameras close to the configuration during data collection and reuse the trained CNN repeatedly.

While this supervised approach still requires a motion capture setup to initially collect the ground truth data for training, the user can realign their cameras to perform inference using the original training data. Given our realignment utility, the cameras can be reset to the approximate same relative poses to the robot's base, then the trained model can successfully estimate the key point positions of the robot without requiring a motion capture system to retrain.

\section{Results}

\subsection{Experimental Setup}
\subsubsection{Soft Robots for Evaluation}
The approach is tested on two types of soft robotic arms\cite{katzschmann2019dynamic, toshimitsu2021sopra} and a soft robotic fish\cite{zhang2022creation} (\textbf{Figure 3}).
The first soft robotic arm (which we shall refer to as the \textit{WaxCast} arm) consists of three axially connected cylindrical segments, each with four separately inflatable chambers.
They are inflated using air provided through a pressure-controlled valve array (\textit{Festo SE \& Co. KG}). By inflating one side, the chambers on that side elongate and induce bending in the segment, thus, the bending direction of the arm can be chosen by selecting the corresponding combination of inflation chambers.
Each segment has a length of about 110 $\pm$ 1 mm and a diameter of 40 $\pm$ 1 mm.
The combined length of the arm is 335 $\pm$ 3 mm. The second arm, \textit{SoPrA}, is a two-segment soft robotic arm with fiber-reinforced pneumatic actuators.
Segments are made of three individually fiber-reinforced elastomer air chambers that are glued together. Combining two of these segments adds up to a total length of 270 $\pm$ 2 mm.
The robotic fish tail is similar in construction and actuation compared to the \textit{WaxCast} arm, except that it is shaped like a fishtail.
It consists of two inflatable chambers and has a total length of 115 $\pm$ 1 mm.

\begin{figure}
    \centering
    \includegraphics[width=\columnwidth]{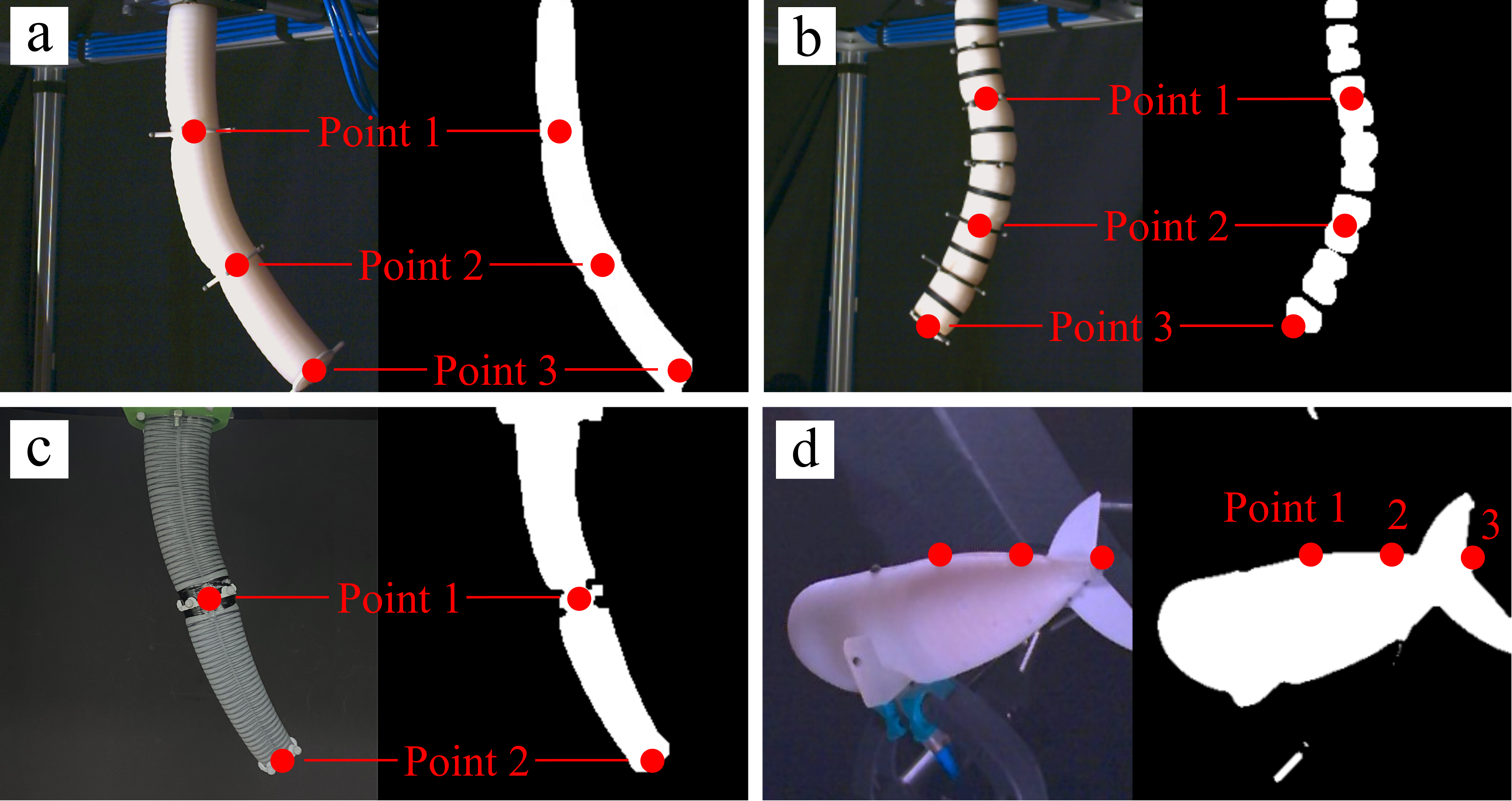}
    \caption{Soft robots used for performance evaluation with evaluation point positions illustrated, in each panel, left shows the original RGB image, right shows the preprocessed image. (a) \textit{WaxCast} Arm,\cite{katzschmann2019dynamic} (b) \textit{WaxCast} arm with visual features (black stripes),\cite{katzschmann2019dynamic} (c) \textit{SoPrA} arm,\cite{toshimitsu2021sopra} (d) Soft fish.\cite{zhang2022creation}}
    \label{fig:soft_robots}
\end{figure}

\subsubsection{Data Collection}
\label{sec:data_collection}
The ground truth data for learning is obtained by eight \textit{Miqus M3} motion-capture cameras from \textit{Qualisys AB} placed in the motion capture space of 1.6 $\times$ 1.1$\times$ 0.8 m.
The placement of the motion-capture markers can be seen in Figure 3.
A group of reflective markers is placed on a rigid ring at the end of every segment of the soft arms.
Along the soft fish's tail, the markers are placed with spaces of 38 $\pm$ 1 mm between them.
Marker position data is supplied at 100 Hz with an average accuracy of 0.1 $\pm$ 0.2 mm, while RGB image data is recorded at 25 Hz by two depth cameras (\textit{Intel RealSense D435i}).\cite{keselman2017intel}
During the data acquisition, the pose data (\textit{i.e.}, the specific shape configurations of the soft robots) is collected such as to cover the robots' full workspace.
The segments of the \textit{WaxCast} arm are all actuated to perform a circular motion, with periods of 100, 10, and 1 second, for segments 1, 2, and 3, respectively. 
We created two data sets, one with three motion-capture marker rings on the arm and the other with six, which also contain visual features in the form of black stripes that were put on the arm (Figure 3b).
This process was repeated for \textit{SoPrA}, but with the chambers randomly actuated.
In total, three labeled data sets are generated for two arms, each containing 12,000 poses.
The soft robotic fish is actuated to perform a tail-fin stroke with maximal deflection, resulting in a data set of 1,800 poses.
Each data set is split into 90\% training and 10\% testing sets.

\subsubsection{Network training}
\label{sec:network_training}
The network is implemented and trained using the \textit{PyTorch} framework.
\textit{AdamW} is chosen as an optimizer and used to minimize the mean absolute loss.\cite{loshchilov2017decoupled}
The network is trained per robot using a batch size of 64 with an early stop for a maximum of 450 epochs on each data set.
The learning rate is set to $10^{-4}$ and reduced by 0.5 after each 200th epoch.
Dropout is applied with a probability of 0.5 in the fully connected layers during training to avoid overfitting.
Training on a GPU (Nvidia GeForce RTX 3090) requires between 30 to 60 minutes to converge.

\subsection{Parametric Shape Representations}
\label{sec:shape_representations}
The CNN was trained using the image data and either learned to output parameters of a PCC model that was fitted to the ground truth marker data or virtual marker positions along the arm (point estimation).
We also analyze the approach's accuracy when estimating just three PCC sections or virtual points compared to estimating six PCC sections or points.
Both PCC and point estimation approaches were tested using the data sets from our two \textit{WaxCast} arms (\textbf{Figure 4a-d}).
Detailed results of the evaluation using VGG-s-bn can be seen in Table 1 for Experiment a-d, with the point estimation approach strictly outperforming the PCC approach.
The errors are normalized based on the robot's length, which is 335 $\pm$ 3 mm for the soft arm. 

\begin{table}[h]
    \centering
    \caption{Estimation errors of the piecewise constant curvature (PCC) and point estimation approaches.}
    \resizebox{\columnwidth}{!}{ 
    \begin{tabular}{ c  c  c  c  c  c  c }
        \hline
        Exp. & Repr.-Sec. & Out.\textsuperscript{a)} & Feat.\textsuperscript{b)} 
        & Point 1 [\%]\textsuperscript{c)} & Point 2 [\%] & Point 3 (tip) [\%]\\
        \hline
        a & PCC-6 & 12 & Yes & 0.9 $\pm$ 0.3 & 3.2 $\pm$ 1.4 & 6.1 $\pm$ 2.4\\
        b & PCC-3 & 6  & No  & 1.3 $\pm$ 0.6 & 2.6 $\pm$ 1.3 & 6.8 $\pm$ 3.5\\
        c & Point-6 & 18 & Yes & 0.06 $\pm$ 0.03 & 0.1 $\pm$ 0.07 & \textbf{0.4 $\pm$ 0.3}\\
        d & Point-3 & 9  & No  & 0.4 $\pm$ 0.4 & 0.8 $\pm$ 1.1 & 3.6 $\pm$ 5.0\\
        e & Point-3 & 9  & Yes & 0.06 $\pm$ 0.03 & 0.1 $\pm$ 0.06 & \textbf{0.3 $\pm$ 0.2}\\
        \hline
    \end{tabular}
    } 
    \vspace{0.5em}
    
    \raggedright
    \textsuperscript{a)}CNN output size \quad
    \textsuperscript{b)}Visual features \quad
    \textsuperscript{c)}Point positions in Figure 3, distance error normalized with the robot's length (335 $\pm$ 3 mm)
\end{table}

\begin{figure*}
    \includegraphics[width=2\columnwidth]{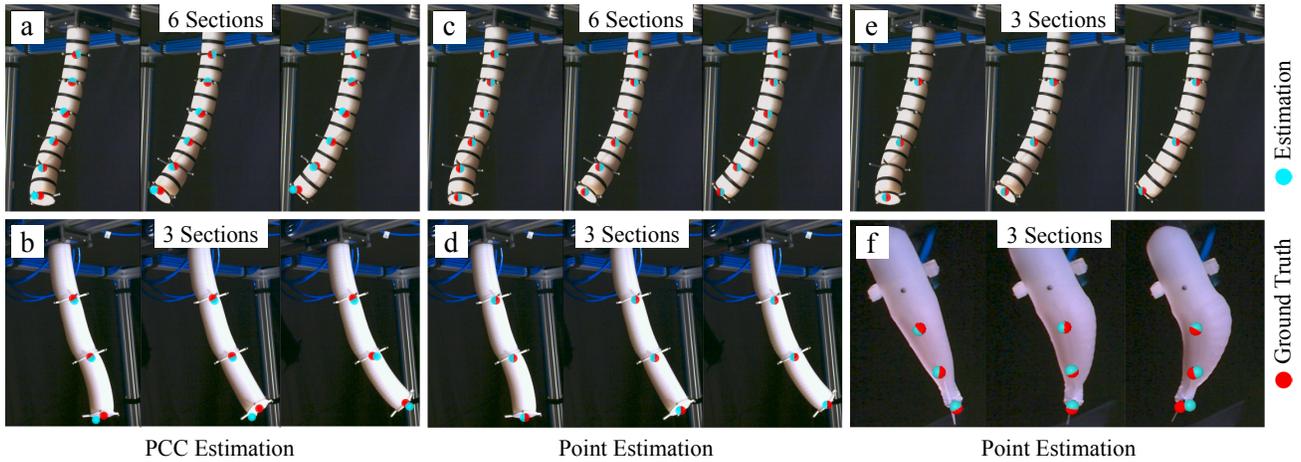}
    \caption{Estimation results of VOKE compared to ground truth positions. Experiment in (a) and (b) employ piecewise-constant curvature (PCC) model, while Experiment (c) to (e) estimate the positions of characteristic points separately. The number of sections considered in each experiment is shown in the figure. The red dots mark the ground truth positions obtained by the motion capture system and the blue dots mark the position estimated by VOKE.}
    \label{fig:result}
\end{figure*}

\begin{table*}
    \centering
    \caption{Performance comparison of different CNN architectures and sizes: VGG, ResNet, EffNet, EffNetV2. Best performance per column is shown in \textbf{bold}. Poor performance values are indicated in ITALICS.}
    \begin{tabular}{ c  c  c  c  c  c  c  c  c  c  c  c }
    \hline
    & 
    &
    & \multicolumn{3}{c}{WaxCast Arm (335 mm)}
    & \multicolumn{3}{c}{SoPrA (270 mm)}
    & \multicolumn{3}{c}{Soft Fish (115 mm)} \\
    %
    & Params\textsuperscript{a)}
    & Forward\textsuperscript{b)}
    & Test\textsuperscript{c)}
    & Max\textsuperscript{d)}
    & Overfit\textsuperscript{e)}
    & Test
    & Max
    & Overfit
    & Test
    & Max
    & Overfit \\
    & [MB]
    & Freq [Hz]
    & [\%]
    & [\%]
    & Ratio
    & [\%]
    & [\%]
    & Ratio
    & [\%]
    & [\%]
    & Ratio
    \\
    \hline
    VGG-s-bn &
    8.30 & 
    \textbf{896}&
    \textbf{0.30} &
    \textbf{2.61} &
    1.06 &
    1.05 &
    4.17 &
    1.15 &
    2.04 &
    9.5 &
    1.13 
    \\

    VGG11 &
    129 & 
    818&
    0.47 &
    9.57 &
    1.22 &
    0.56 &
    \textbf{4.06} &
    1.42 &
    0.84 &
    9.89 &
    \textit{3.29}
    \\

    VGG11-bn &
    129 & 
    561&
    2.86 &
    10.07 &
    1.00 &
    6.89 &
    28.62 &
    1.00 &
    5.17 &
    28.81 &
    0.88 
    \\
    
    VGG13 &
    128 & 
    888 &
    0.44 &
    8.60 &
    1.35 &
    \textbf{0.54} &
    4.22 &
    1.44 &
    0.82 &
    12.19 &
    \textit{3.23}
    \\

    VGG13-bn &
    129 & 
    578 &
    3.00 &
    13.32 &
    0.99 &
    6.73 &
    29.83 &
    1.00 &
    4.83 &
    26.00 &
    0.87 
    \\

    ResNet18 &
    11.2 & 
    395&
    0.75 &
    6.81 &
    1.44 &
    0.95 &
    6.60 &
    1.19 &
    0.70 &
    4.81 &
    \textit{1.83}
    \\

    ResNet34 &
    21.3 & 
    188&
    0.72 &
    4.91 &
    \textit{1.54} &
    0.91 &
    23.36 &
    1.15 &
    0.64 &
    4.75 &
    \textit{2.06}
    \\

    ResNet50 &
    23.5 & 
    126&
    0.75 &
    6.23 &
    \textit{3.55} &
    0.65 &
    17.34 &
    \textit{1.59} &
    0.64 &
    4.24 &
    \textit{3.11} 
    \\

    EffNet-B0 &
    \textbf{4.01} & 
    75&
    1.99 &
    13.35 &
    \textit{1.70} &
    0.89 &
    8.42 &
    1.20 &
    0.75 &
    16.13 &
    \textit{2.25}
    \\

    EffNet-B1 &
    6.52 & 
    62&
    1.57 &
    10.27 &
    1.43 &
    1.02 &
    6.28 &
    1.10 &
    0.73 &
    5.83 &
    \textit{1.91}
    \\

    EffNet-B3 &
    10.7 & 
    51&
    1.65 &
    9.58 &
    \textit{1.61} &
    1.01 &
    7.15 &
    1.10 &
    0.71 &
    6.90 &
    \textit{1.77}
    \\

    EffNet-B5 &
    28.4 & 
    38&
    1.22 &
    8.68 &
    1.24 &
    0.87 &
    5.19 &
    1.09 &
    0.69 &
    5.90 &
    \textit{1.69}
    \\

    EffNetV2-s &
    20.2 & 
    51&
    0.88 &
    11.67 &
    1.16 &
    0.77 &
    4.96 &
    1.04 &
    0.59 &
    3.79 &
    \textit{1.58} 
    \\

    EffNetV2-m &
    52.9 & 
    29&
    0.75 &
    17.47 &
    1.00 &
    0.66 &
    13.52 &
    1.04 &
    \textbf{0.62} &
    \textbf{5.32} &
    1.39
    \\
    \hline
\end{tabular}
\newline
\textsuperscript{a)}Parameters with output size 6
\textsuperscript{b)}Averaged over 3 soft robots
\textsuperscript{c)}Mean tip estimation error of testing set, normalized with the robot's length
\textsuperscript{d)}Maximum tip estimation error of testing set, normalized with the robot's length
\textsuperscript{e)}Test over Train mean tip estimation error, high overfit ratios are marked in italics
\vspace{-1em}
\end{table*}

\subsection{Visual Features}
\label{sec:visual_features}
To evaluate the effect of visual features on the estimation accuracy, the point estimation approach is applied to a data set with features (Figure 4e).
We modified the \textit{WaxCast} arm's appearance to have multiple black stripes perpendicular to the arm's backbone (Figure 3b).
In Table 1, Experiment d and e show that the mean tip error for the feature-less \textit{WaxCast} arm is 3.6 $\pm$ 5.0\% and only 0.3 $\pm$ 0.2\% for the arm with features.

\subsection{CNN Architectures}
We tested 14 different CNN architectures and reported their performance on three soft robots in \textbf{Table 2}.
All errors are normalized with the corresponding robot's length.
The main performance metrics are mean and maximum tip estimation errors of the testing data set. Considering the limited data size, we prefer networks with better generalizability outside of the training data set.
Therefore, we also include the overfit ratio, which is calculated by dividing the mean tip estimation error of the testing data set by the training mean errors.
To avoid possible overfitting to the training set, for each data set, the best CNN performance is picked among architectures with the overfit ratio under 1.5. 
Above this threshold, the testing error is more than 1.5 times the training error, suggesting an unbalanced performance not generalizable throughout the whole workspace.
The best estimation performance for the \textit{WaxCast} arm is 0.3\% (1.01 mm) by VGG-s-bn, for \textit{SoPrA}, it is 0.54\% (1.46 mm) by VGG13, and for the soft robotic fish, it is 0.62\% (0.72 mm) by EffNetV2-m.
We also report the parameter numbers with output size 6 and the average CNN forward frequency for a single estimation on a cluster GPU (NVIDIA Tesla V100-SXM2 32 GiB).

\subsection{Maximum Estimation Errors}
\label{sec:max_estimation_errors}
The maximum estimation errors are computed as an indication of the “worst case scenarios”.
For \textit{WaxCast} arm with features, the maximum tip error is 2.6\% (VGG-s-bn), for \textit{SoPrA}, it is 4.2\% (VGG13), and for the soft robotic fish, it is 5.3\% (EffNet-B0).

\begin{table*}
    \caption{
    Estimation errors of our approach compared to other works.
    }
    \begin{tabular}{ c  c  c  c  c  c  c  c  c}
    \hline
    %
     & Estimation
    & \# of
    & Require
    & Shape
    & Robot
    & Robot
    & Tip Error\textsuperscript{a)}
    & Freq.\textsuperscript{c)}
    \\
    & Technique
    & Cams
    & Exact
    & Agnostic
    & Type
    & Length
    & [\%]
    & [Hz]
    \\

    &
    &
    & Contour
    &
    &
    &[mm]
    &
    \\

    \hline
    
    \textbf{Ours (VGG-s-bn)} &
    CNN & 
    2 & 
    \textbf{No} &
    \textbf{Yes} &
    WaxCast arm &
    335 &
    \textbf{0.3 $\pm$ 0.2} &
    146
    \\

    \textbf{Ours (VGG13)} &
    CNN & 
    2 & 
    \textbf{No} &
    \textbf{Yes} &
    SoPrA &
    270 &
    \textbf{0.5 $\pm$ 0.4}&
    106
    \\

    \textbf{Ours (EffNetV2-m)} &
    CNN & 
    2 & 
    \textbf{No} &
    \textbf{Yes} &
    Soft fish &
    115 &
    \textbf{0.6 $\pm$ 0.6} &
    8.6
    \\

    Camarillo et al.\cite{camarillo2008vision} &
    2D point- & 
    3 & 
    Yes &
    No &
    Soft arm &
    160 &
    4.8 &
    3\textendash4
    \\
    
    &
    cloud fit &
    &
    &
    &
    &
    &
    \\

    Vandini et al.\cite{vandini2017unified} &
    Line feature & 
    1 & 
    \textbf{No} &
    No &
    Soft arm &
    260 &
    2.8 &
    0.1
    \\
    
    &
    detector &
    &
    &
    &
    &
    &
    \\
    
    Pedari et al.\cite{pedari2019spatial} &
    LED light & 
    2 & 
    \textbf{No} &
    \textbf{Yes} &
    Soft arm &
    468\textsuperscript{b)} &
    4.5 &
    N/A
    \\
    
    &
    placement &
    &
    &
    &
    &
    &
    \\
    
    AlBeladi et al.\cite{albeladi2021vision} &
    Edge detector & 
    1 & 
    Yes &
    No &
    Soft arm &
    287 &
    4.5 $\pm$ 3.1 &
    N/A
    \\
    
    &
    \& curve fit &
    &
    &
    &
    &
    &
    \\
    \hline
\end{tabular}
\newline
\textsuperscript{a)} Error normalized with the corresponding robot's length
\textsuperscript{b)} Not provided, calculated based on their estimation data
\textsuperscript{c)} Estimation frequency as reported in original works, not tested on the same machine, our results are tested with a NVIDIA GeForce 3090 24 GiB GPU.
\\
    \label{tab:comparison}
\end{table*}

\subsection{Benchmarks}
\label{sec:benchmarks}
We compared the results of our point estimation approach with four similar works (see \textbf{Table 3}), which also estimated reference points along continuously deforming shapes. Since the previous works either are shape agnostic or employ active LED markers,\cite{albeladi2021vision,camarillo2008vision,vandini2017unified,pedari2019spatial} they are unsuitable for our application with different soft robots.
Due to limited data and the lack of open-source code/platform to reproduce the benchmarks, especially for the concentric robots for medical use, we only compared the tip errors, which are usually the largest and normalized with the length of each corresponding robot for a fair comparison.
We believe that achieving low tip position reconstruction error is important for soft robotic shape estimation methods since this accuracy is critical in real-world operations involving reaching and grasping of objects.

\textit{DeepLabCut} is not included in Table 3 because it estimates pixel locations instead of 3D positions.\cite{mathis2018deeplabcut}
To compare the results, we projected the estimated and ground truth 3D positions into the input images and evaluated the pixel distance error.
Experiment c in Table 3 showed a root-mean-square error at the tip position of 1.13 pixels in one camera view and 1.18 pixels in the other. In comparison, \textit{DeepLabCut} achieved an accuracy similar to the human labeling error of 2.69 $\pm$ 0.1 pixels.
However, comparing pixel errors is only of limited value, since a pixel error can have a different significance depending on the image resolution and scale of the captured object.
Moreover, reprojecting the estimated pixels from multiple calibrated cameras back into 3D space may bring in additional errors due to camera calibrations.
Therefore we believe that a direct 3D position estimation is more useful and convenient for downstream applications.

\subsection{Online Estimation}
\label{sec:online_estimation}
The online estimation performance of VOKE was tested on both a portable computer (2-core, 2.70 GHz Intel Core i7-7500U CPU, no GPU) and an Omen desktop computer (24-core, 3.2GHz Intel Core i9-12900K CPU, 64GB memory, NVIDIA GeForce 3090 GPU with 24GB memory).
A single estimation using our truncated VGG-s-bn architecture for SoPrA takes 54 ms (18.4 Hz) on average on the portable computer, of which 60\% are used for the CNN forward calculation and 38\% are used for image processing.
The remaining 2\% are used to stream the images from the RGB cameras. The estimation rate can be greatly improved with the use of a GPU. CNN forward calculation and image processing on the desktop computer take only 1.96 ms and 6.77 ms, respectively, giving a theoretical estimation frequency of over 100 Hz. However, the real-world online estimation rate in this case is currently limited by the camera frequency of 30 Hz.

\subsection{Different Experimental Setups}
\label{sec:diff_exp_setups}
To demonstrate the robust performance of our proposed pipeline with adaptive thresholding in a range of lighting conditions without the need of retraining CNN, we evaluate and present the tip estimation errors on the \textit{SoPrA} testing data set with modified brightness levels (see \textbf{Figure 5a}) and added Gaussian noise (Figure 5b).
The brightness of the original images is modified by adding or subtracting pixel values from the grayscale images (pixel value range 0-255). 
The Gaussian noise is added per pixel to the original gray-scale images with increasing standard deviation of the noise distribution.
The experiments are conducted on the \textit{SoPrA} data set with the best-performed VGG13 network since the gray color of the \textit{SoPrA} arm is the closest to the black background.
\textit{SoPrA} provides the least contrast compared to the other soft robots and is, therefore, more sensitive to changes in brightness.

\begin{figure*}
    \includegraphics[width=2\columnwidth]{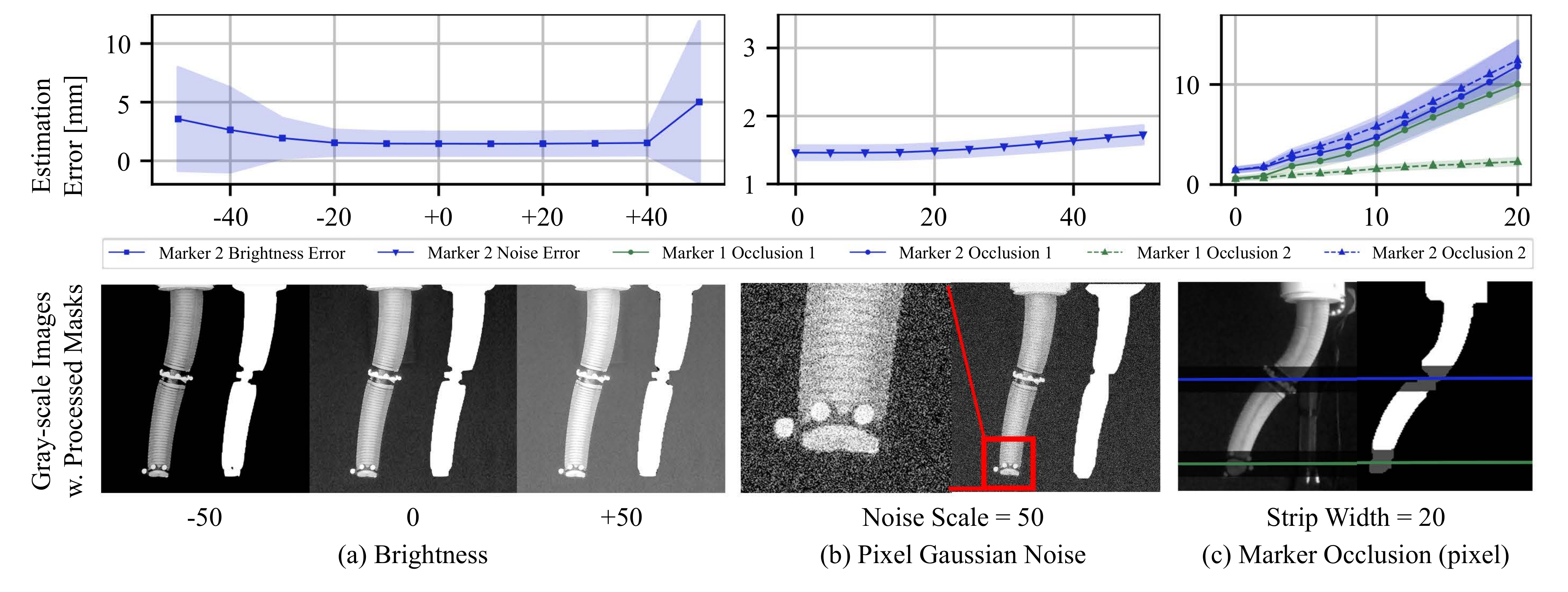}
    \caption{Performance of VGG13 on \textit{SoPrA} under different experimental setups, with sample gray-scale images and processed masks shown. (a) Tip estimation errors with varying image brightness. The brightness modification is quantified by the addition or subtraction of pixel values (0\textendash255) from original gray-scale images. (b) Tip estimation errors with increasing Gaussian noise. The noise with standard deviation from 0 to 50 is added per pixel to the pixel values of original gray-scale images. (c) Marker 1 and marker 2 (tip) estimation errors plotted with black strip occlusions of varying width at the marker positions.}
    \label{fig:lighting}
\end{figure*}

In the scope of this work, we do not explicitly consider the problem of occlusion. Preliminary evaluation is carried out by inserting black strips of varying width at the marker position to test the inference robustness of the trained VGG13 network. The result is shown in Figure 5c.

The performance of the trained VGG13 network is also tested after the reassembly of the cameras.
With relative camera translations and orientations obtained from the fiducial markers, we manually realign the cameras to a configuration with 1.46 mm difference from the one used during data collection.
The new tip estimation error after reassembly of the cameras is 1.5 $\pm$ 1.6\% for the \textit{SoPrA} test data set.

\subsection{Discussion}
\label{sec:discussion}
The results show that the estimation errors increase along the shape regardless of the data set or shape representation being used.
This increase is most likely due to the fact that the tips of these shapes typically move faster and across a larger space than the rest of their shapes. A dynamic behavior increases the estimation difficulty towards the tip.

The approach using the PCC shape representation as output produces larger estimation errors on the three tested data sets (Table 1).
This error is partially because the endpoint positions are computed using forward kinematics calculation with all previous PCC sections, accumulating the estimation errors of each section.
Another reason for the inferior performance of PCC is that the constant curvature assumption is sometimes inaccurate for a real soft robotic system.
For example, the sections of the soft arm do not exactly bend with constant curvature.
The arm's weight and dynamics, the design characteristics of the inflation chambers, and the fabrication errors all introduce imperfections with regard to the constant curvature assumption.
Moreover, the arms we use in this work do not contain an inextensible backbone and therefore also extend along their center line under inflation.
This limitation could be resolved by augmenting the PCC model to allow for constant curvature sections of variable length.
The CNN would then need to be adjusted to also estimate the length of each segment.
However, the error due to non-constant curvature deformation would remain.
By estimating points separately, we could avoid both the error accumulation and the PCC model limitations.
Although the point representation does not contain any statements about connectedness or directionality, it gives a more precise estimation of the tip position.

The visual features added to the \textit{WaxCast} arm greatly improved the estimation accuracy.
This can be seen when comparing Experiments d and e in Table 1.
We believe that the added features helped the CNN extract more information from the input images.
The increased information content improved the deduction of the shape parameters.

Performance of different CNN architectures is presented in Table 2.
Overall, VGG architectures exhibit the fastest estimation speed and the least tendency to overfit the training data set. Although the performance of VGG with batch normalization greatly decreased, we can see that the batch normalization helps with preventing overfitting.
For the larger data sets of the \textit{WaxCast} and \textit{SoPrA} arms, VGG and EfficientNetV2 architectures do not overfit the training sets.
However, due to the limited data set size for the soft fish, most CNN architectures (other than VGG with batch normalization and EffNetV2-m) tend to overfit for this training set. 
This overfitting is also because more recent CNN architectures, especially EfficientNet and EfficientNetV2 are designed to scale up training with larger image sizes, therefore they might overfit small data sets more easily and cannot outperform simpler VGG architectures on tasks with small binary image input.
Among the architectures that do not overfit on the training data of the soft fish, EffNetV2-m performs the best in terms of the mean estimation error on the testing set. In practice, the EffNetV2-m architecture is least favored for our application since its average online estimation frequency is below 10 Hz.

We outperform the benchmarks for all three soft robots as shown in Table 3.
At the same time, our approach also does not require to extract contour lines or any prior knowledge of the shape, suggesting the possible generalizability to different types of soft robots.

One limitation of using convolutional neural networks is that the trained network may not be reusable and needs retraining when the experimental setup changes.
We tested our trained network on input images with various levels of brightness, noises, occlusion, and after reassembly of the cameras.
The stable performance under brightness changes and gaussian noises (Figure 5a and 5b) indicates that the proposed method could work with a wide range of lighting conditions without re-training the CNN as long as there is sufficient contrast for the adaptive image preprocessing.
Since we do not consider occlusions during the training phase, the estimation performance decreases with increasing occlusions during inference. However, compared to marker-based method, the CNN is capable of predicting marker positions even with the markers fully occluded. When occluding marker 2 up to a width of 20 pixels, the estimated position error of marker 1 only slightly decreases by 
0.63 $\pm$ 0.39\%
compared to the unoccluded case (Figure 5c). This shows the potential robustness of the proposed method against occlusion.
Although retraining would be needed for different camera configurations, we show with the aid of fiducial markers (\textit{AprilTags}), rough realignment to previous camera positions is possible and the trained network can be reused.

\section{Conclusion}
\label{sec:conclusions}
VOKE is a vision-based, 3D soft robot key point estimation approach using two cameras and a CNN.
It outperforms current marker-less estimation approaches when evaluated on two soft robotic arms and one soft robotic fish.
While we consider the visual robustness of our approach to be an improvement over the state-of-the-art, it could be further enhanced to be calibration-free, deal with occlusions, and allow for more expressive representations.
Future work will introduce artificial occlusions in the network's training process to work with partially occluded images and also employ learning-based shape segmentation to perform robust background removal under insufficient contrast. Another future direction is to generalize the approach to the estimation of more expressive kinds of shape representations, \textit{e.g.}, mesh reconstructions, instead of being limited to the estimation of piecewise constant curvatures or characteristic points.








\bibliographystyle{./IEEEtranBST/IEEEtran}
\bibliography{./IEEEtranBST/IEEEabrv,./refs}

\end{document}